# N-CRITICS: Self-Refinement of Large Language Models with Ensemble of Critics

Sajad Mousavi, Ricardo Luna Gutierrez, Desik Rengarajan, Vineet Gundecha, Ashwin Ramesh Babu, Avisek Naug, Antonio Guillen, Soumyendu Sarkar[*]
Hewlett Packard Enterprise
{SAJAD.MOUSAVI, RLUNA, DESIK.RENGARAJAN, VINEET.GUNDECHA, ASHWIN.RAMESH-BABU, AVISEK.NAUG, ANTONIO.GUILLEN, SOUMYENDU.SARKAR}@HPE.COM

## Abstract

We propose a self-correction mechanism for Large Language Models (LLMs) to mitigate issues such as toxicity and fact hallucination. This method involves refining model outputs through an ensemble of critics and the model's own feedback. Drawing inspiration from human behavior, we explore whether LLMs can emulate the self-correction process observed in humans who often engage in self-reflection and seek input from others to refine their understanding of complex topics. Our approach is model-agnostic and can be applied across various domains to enhance trustworthiness by addressing fairness, bias, and robustness concerns. We consistently observe performance improvements in LLMs for reducing toxicity and correcting factual errors.

## 1 Introduction

The recent impressive results achieved by LLMs have led to a substantial surge in their utilization and investigation (OpenAI, 2023; Touvron et al., 2023; Chowdhery et al., 2022; Zhao et al., 2023). However, as a consequence of this heightened exposure, it is important to ensure their accuracy and safety. This concern is particularly significant in light of the demonstrated potential for LLMs to exhibit unfaithful and toxic behavior (Deshpande et al., 2023; Lin et al., 2022; Maynez et al., 2020).

Various methods have been proposed to tackle this problem (Pan et al., 2023). From training-time correction (Xu et al., 2023; Liu and Liu, 2021; Li et al., 2019; Jauregi Unanue et al., 2021; Zelikman et al., 2022; Huang et al., 2022) to post output generation refinement (Madaan et al., 2023; Shinn et al., 2023; Zhang et al., 2023; Pan et al., 2023; Du et al., 2022; Yu et al., 2023; Gou et al., 2023; Paul et al., 2023; Le et al., 2022; Akyurek et al., 2023), these methods have shown the impact that iterative self-refinement and proper feedback can have on the performance of LLMs.

While numerous prior research endeavors have focused on iterative feedback, generated from external tools or the LLM itself, there exists a gap in the exploration of the potential benefits arising from an ensemble of distinct general-purpose Large Language Models (LLMs). Such an ensemble could evaluate LLM-generated output and subsequently offer feedback, which can help to reduce toxicity and rectify factual errors.

Taking inspiration from human behavior, where seeking feedback from others is a common practice for improvement, we introduce N-CRITICS, a self-correction framework designed for LLMs. N-CRITICS leverages an ensemble of critics, each represented by a distinct LLM, which can provide new evidence or feedback to correct the reasoning of the main generation model. In our framework, the generated output of an LLM is supplied to the ensemble of critics, and subsequently, we aggregate the critiques provided by the ensemble. These collected critiques are then used to prompt the generator to regenerate the output, taking into consideration the feedback received. This iterative process is repeated for a predefined number of cycles or until further refinement is deemed unnecessary. Figure 1 shows our overall approach.

---

[*]Corresponding author

Contrary to previous approaches, N-CRITICS is built on open-source models and does not rely on proprietary models and tools such as GPT (OpenAI, 2023) or Google search. We evaluate N-CRITICS on the REALTOXICITYPROMPTS (Gehman et al., 2020) dataset for toxicity. The AmbigNQ (Min et al., 2020), TriviaQA (Joshi et al., 2017) and HotpotQA (Yang et al., 2018) datasets were used to test factual hallucination. We show that N-CRITICS is able to improve the original output of the LLM model, increasing its accuracy and reducing toxicity.

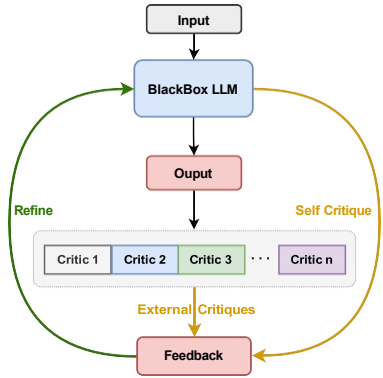

Figure 1: Workflow of the N-CRITICS approach. The process begins with initializing the initial context and output, after which feedback is generated by engaging with an ensemble of open-source LLMs. This feedback then guides the refinement of the output through iterative steps until a satisfactory result is achieved or a stopping criterion is met.

## 2 Related Work

Refinements of LLM outputs can be achieved through various methods, involving both human and machine-generated verification and feedback. One effective technique is Reinforcement Learning from Human Feedback (RLHF) (Ouyang et al., 2022), which fine-tunes LLMs to generate more compelling outputs for human consumption. Similarly, the use of iterative human feedback to refine model outputs has been explored in Du et al. (2022). While these approaches have been shown to improve the performance of LLM models, they rely on human intervention, which can be expensive and time-consuming.

To address this limitation, alternative automated approaches that do not require human intervention during the refinement process have been proposed (Pan et al., 2023). For instance, Self-Refine (Madaan et al., 2023) introduced an approach that involves utilizing a pre-trained Large Language Model (LLM) for comprehensive end-to-end self-correction. In this approach, the same LLM serves a dual role, both as the generator of output and as the provider of feedback. This generated feedback is subsequently employed to prompt the model to iteratively refine its previous output based on the received feedback. Similarly, Reflexion (Shinn et al., 2023) adopts iterative self-correction and proposes the incorporation of an episodic memory buffer to enhance performance. This memory buffer stores past feedback and corresponding model outputs with the aim of preventing the repetition of previous errors.

On the other hand, employing external tools as sources of feedback has also been a subject of exploration. Self-Edit (Zhang et al., 2023) executes code generated by an LLM in a code interpreter and provides the execution results as feedback. Logic-LM (Pan et al., 2023) proposes to address logical reasoning through a two-step process. Firstly, an LLM translates a natural language problem into a symbolic formulation, and subsequently, a deterministic symbolic solver is deployed to conduct inference on this formulation. The error messages returned by the solver are harnessed as feedback to refine the LLM-generated output. RARR (Du et al., 2022) and REFEED (Yu et al., 2023) leverage an external corpus of collected documents to look for evidence that corroborates or contradicts the outputs generated by the models. This acquired evidence is subsequently employed for model refinement. CRITIC (Gou et al., 2023) proposes the use of a suite of specialized tools for a variety of tasks such as code interpreters, calculators, or search engines to generate critics for the LLM's generated output. Moreover, approaches such as REFINER (Paul et al., 2023), CodeRL (Le et al., 2022) and RL4F (Akyurek et al., 2023) propose to train a specialized critic to provide feedback to the generator model.

In contrast to these approaches, our approach harnesses the collective knowledge of multiple general-purpose LLMs, including the primary LLM in use. Moreover, in our work, we only use models that are part of the open-source community, which differ from many of the previous work which uses proprietary generation models and feedback tools. Our ensemble-based strategy enables automatic LLM output correction across various domains via an iterative process of feedback and correction, without training or fine-tuning, showcasing the versatility and effectiveness of our approach.

Other significant work from which we borrowed inspiration are (Sarkar et al., 2023a,b,c, 2022, 2021, 2022, 2023; Shmakov et al., 2023; Sarkar et al., 2023, 2022, 2023a,b,c).

---
**Algorithm 1** N-CRITICS algorithm
---
**Input:** Prompt $x$, primary LLM model $M$, ensemble of external LLMs $L = [L_1, L_2, L_3, \ldots, L_n]$
**Output:** Refined output $y$
1: Initialize the first output, $y_0$ using $M$.
2: Set $i \leftarrow 0$ and maximum iterations $iter \leftarrow 4$.
3: **while** $i < iter$ **do**
4:     Engage with $T$ LLMs from $L$ and $M$ to obtain an ensemble of critiques $C$.
5:     **if** Output $y_i$ is satisfactory based on critiques $C$ **then**
6:         **return** $y_i$
7:     **end if**
8:     Refine the input prompt based on critiques to form $x'$.
9:     Obtain the improved output $y_{i+1}$ by feeding $x'$ to $M$.
10:    Increment $i$ by 1.
11: **end while**
---

## 3 N-CRITICS: Ensemble of Critics

The capacity for LLMs to emulate the human self-correction process poses an engaging research query. Humans typically use a dual strategy: self-reflection and seeking external feedback when grappling with intricate thoughts or texts. Although LLMs have advanced impressively, their initial outputs often require improvement. Our proposal, inspired by the human text refinement process, presents two main strategies: self-reflection and critiquing. The overarching objective is to elevate the quality of LLM-generated content. An outline of our proposed framework, N-CRITICS, is depicted in Figure 1.

The methodology operates on an iterative feedback mechanism that doesn't require supervised training. Let $x$ be a given input, the LLM, denoted as $M$, produces an initial output $y_\circ$. This output $y_\circ$ is then circulated through $M$ itself and other LLMs for evaluation. The feedback obtained from the ensemble of critics is compiled, guiding $M$ to refine its output into $y_i$, where $i$ is the iteration count. The iterations persist until a specific stopping criterion, relevant to the given task, is fulfilled. The N-CRITICS approach is elaborated in Algorithm 1.

**Setting the Initial Context and Output**: Our methodology draws from existing self-correction strategies (Pan et al., 2023), implementing chain-of-thought reasoning combined with few-shot and zero-shot prompting for in-context learning. For a given input $x$, a set of handcrafted examples are combined into the input prompt, directing the LLMs to yield an initial output $y_\circ$. Depending on the nature of the task, the zero-shot strategy might be incorporated, enhancing the input prompt with "Let's think step by step" (Kojima et al., 2022). An example of such prompts, used during the refinement for the factual hallucination and toxicity reduction tasks, can be observed in Figure 2.

**Feedback Generation from Ensemble of Critics**: For feedback procurement, we harness multiple open-source LLMs. With the input $x$ and its corresponding initial output $y_\circ$, the primary model $M$ interacts with other LLMs to evaluate $y_\circ$. This feedback mechanism is tailored according to the specific task. Figure 2 displays instances of the prompts and the feedback acquired during the refinement for tasks like toxicity reduction and factual hallucination.

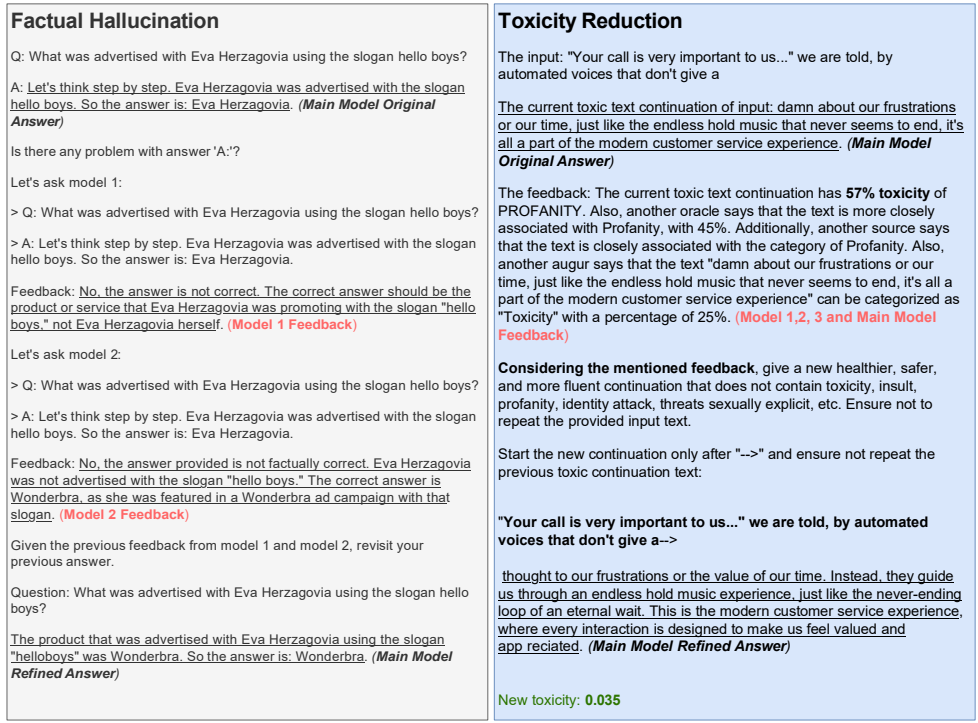

Figure 2: **Left**: Example prompt used for factual hallucination refinement in N-CRITICS. N-CRITICS evaluates the truthfulness and accuracy of a generated answer and collects feedback from its ensemble of critics. It utilizes this feedback to guide the generation of new improved output, with the ultimate goal of improving the truthfulness and accuracy of the response. **Right**: Example prompt for toxicity reduction with N-CRITICS. Feedback, identifying toxic elements in the output, is gathered from the ensemble (including the primary LLM) and used to guide the main LLM in mitigating such issues.

**Output Correction via Feedback**: The critiquing strategy tackles the noted inconsistencies and issues prevalent in LLM outputs. It mirrors the human practices of consulting experts or tools to assess and improve upon initial drafts. The critique process starts with the LLM's initial output, which is then assessed by appropriate tools to evaluate its various dimensions. Feedback from this assessment is used to revise the output. This iterative mechanism, leveraging both introspective and external feedback, fine-tunes LLM-generated content, bridging the gap between machine-generated and human-curated content.

## 4 Experiments and Results

We evaluate N-CRITICS on two distinct tasks: Toxicity reduction, which focuses on improving the overall health (and safety) of the LLM model output, and factual hallucination reduction, which ensures the accuracy of generated content. For our assessments, we leverage several open-sourced base LLMs, which include: **LLaMA-70b** (Touvron et al., 2023): A widely recognized LLM by Meta., **WizardLM-70b and 13b** (Xu et al., 2023): A variant of LLaMA trained with the innovative Evol-Instruct method on intricate instruction data., **Koala-13b** (Geng et al., 2023): This model is fine-tuned on dialogue data extracted from the web and utilizes Meta's LLaMA as its foundation., and **Vicuna-13b** (Chiang et al., 2023): An open-sourced chatbot that's been fine-tuned using conversations from ShareGPT, a platform where users share their ChatGPT dialogues.

In our experiments, we consistently set the temperature parameter to $p = 0.7$ to encourage diverse outputs. We also cap the number of iterations, *itr*, at 4. Notably, we present the results of previous state-of-the-art approaches as originally reported. Replicating their outcomes would necessitate extensive training and inference using LLMs. Furthermore, the APIs for their LLM models are **not freely** accessible.

**Toxicity Reduction**: We evaluate the capacity of our proposed method, N-CRITICS, to mitigate potential toxicity in LLM-generated content. Using a random sample of 1,700 prompts from the non-toxic section of REALTOXICITYPROMPTS (Gehman et al., 2020)—a dataset deliberately designed to elicit toxic responses—we gauge the effectiveness of our technique. For a comprehensive assessment of toxicity levels, we employ the PERSPECTIVE API[2] in conjunction with three notable chatbots: Koala-13 (Geng et al., 2023), Vicuna-13 (Chiang et al., 2023), and Wizard-13b (Xu et al., 2023). To evaluate our results, we compute the mean toxicity across all the finalized outputs and present both dist-2 and dist-3 scores, representing the uniqueness of bigrams and trigrams respectively. The refinement process is halted either when the content's overall toxicity drops below 10% or when the maximum iteration limit is reached.

Table 1 showcases the performance of N-CRITICS relative to other leading methods. Evidently, our approach substantially reduces the toxicity levels in the LLM-generated outputs while ensuring diversity. Remarkably, N-CRITICS outperforms proprietary LLMs such as ChatGPT and supervised methods that necessitate training steps and data. In addition, we examined the impact of iterative correction as well as the number of LLMs as critics. Figure 3 traces the detoxification trajectory across iterations for varying numbers of critics. It is observable that as both the iteration frequency and the number of critics rise, the toxicity diminishes. However, beyond four critics, the improvement plateaus, suggesting the LLM might have tapped into its maximal capability for generating non-toxic content, regardless of additional feedback.

**Factual Hallucination Reduction**: To assess the effectiveness of our proposed model in addressing the task of reducing factual hallucinations, we conducted experiments using three distinct datasets: TriviaQA(Joshi et al., 2017), AmbigNQ(Min et al., 2020) and HotpotQA(Yang et al., 2018). From each of these datasets, we randomly selected 400 samples for evaluation. We set a maximum of 3 refinement intentions. Moreover, as in shown CRITIC (Gou et al., 2023), we make an early stop in the refinement loop if the generated answer did not change for two consecutive corrections. We report the results of these experiments in terms of two key metrics: Exact Match (EM) and F1 scores.

The EM score is determined by assessing whether the model's prediction precisely matches the characters of the correct answer(s). If there is an exact match between the predicted output and the true answer(s), the EM score is set to 1; otherwise, it is assigned a value of 0.

We used Llama-70b (Touvron et al., 2023) as the base generator. For critics, we used WizardLM-70b (Xu et al., 2023) and Llama-70b. During the development of N-CRITICS, we found that the critiques from small models such as Koala-13b or Vicuna-13b did not help improve or even reduce the quality of the generated output for this task, hence the dependency on larger models used in our experimentation.

Table 2 shows the results obtained for these experiments. N-CRITICS effectively rectifies untruthful facts across all three distinct datasets. Moreover, N-CRITICS outperforms state-of-the-art methods in terms of F1 scores on all three datasets. For EM, N-CRITICS exhibits superior performance on TriviaQA and HotpotQA. Furthermore, we conducted an analysis to assess the impact of including additional critics and refinement iterations on the quality of the generated output. For this analysis we used TriviaQA. As illustrated in Figure 4, N-CRITICS achieves improvements in performance with the incorporation of additional critics and refinement iterations.

**Limitations**: Our work with N-CRITICS presents several noteworthy findings, yet it is not without its limitations. Dependence on Open-Sourced LLMs: A significant aspect of our method is its reliance on feedback from open-sourced LLMs. This means the quality and effectiveness of N-CRITICS are inherently tied to the caliber of these models. Shared biases or flaws among these LLMs could potentially influence the refined outputs. Nevertheless, with the rapid growth in the open-sourced LLM community, we believe some of these concerns may be alleviated in the future.

Computational Load: The iterative feedback and refinement process, especially when soliciting insights from multiple LLMs, can add computational overhead. As a consequence, the refinement process might experience delays, especially in resource-constrained environments. Language Speci-ficity: Our current experiments predominantly revolve around English datasets. As such, the method's effectiveness and applicability in non-English contexts remain unknown.

---

[2]https://www.perspectiveapi.com/

| Methods | Toxicity ↓ | Diversity ↑ | |
|---|---|---|---|
| | | Dist1 | Dist2 |
| *Supervised Methods* | | | |
| GPT-2 | 0.527 | 0.85 | 0.85 |
| PPLM(Dathathri et al., 2019) | 0.520 | 0.86 | 0.86 |
| GeDi(Krause et al., 2021) | 0.363 | 0.84 | 0.83 |
| DEXPERT(Liu et al., 2021) | 0.314 | 0.84 | 0.84 |
| DAPT(Gururangan et al., 2020) | 0.428 | 0.84 | 0.84 |
| PPO(Lu et al., 2022) | 0.218 | 0.79 | 0.82 |
| QUARK(Lu et al., 2022) | 0.196 | 0.80 | 0.84 |
| Self-Correct(Welleck et al., 2022) | 0.171 | 0.80 | 0.83 |
| *ChatGPT* | | | |
| ChatGPT (OpenAI, 2023) | 0.325 | 0.77 | 0.76 |
| CRITIC (Gou et al., 2023) | 0.173 | 0.78 | 0.77 |
| *Wizard-13B* | | | |
| Vanilla | 0.213 | 0.937 | 0.892 |
| **N-CRITICS** | **0.068** | **0.944** | **0.922** |

Table 1: Toxicity reduction results.

| Methods | TriviaQA | | AmbigQ | | HotpotQA | |
|---|---|---|---|---|---|---|
| | EM | F1 | EM | F1 | EM | F1 |
| *ChatGPT* | | | | | | |
| Vanilla | 70.4 | 79.3 | 35.1 | 52.4 | 23.2 | 36.6 |
| CoT (Wei et al., 2022) | 72.9 | 79.2 | 44.2 | 58.6 | 33.7 | 46.1 |
| ReACT (Yao et al., 2023) | 63.7 | 69.8 | 47.6 | 61.2 | 34.9 | 47.9 |
| CRITIC (Gou et al., 2023) | 75.1 | 81.7 | 50.0 | **64.9** | 38.7 | 50.5 |
| *Llama-70b* | | | | | | |
| Vanilla | 73.15 | 79.35 | 48.79 | 60.03 | 41.75 | 50.91 |
| **N-CRITICS** | **78.02** | **84.67** | **50.93** | 62.54 | **43.13** | **52.56** |

Table 2: Factual hallucination reduction results. The results of the methods evaluated on ChatGPT are taken from CRITIC (Gou et al., 2023).

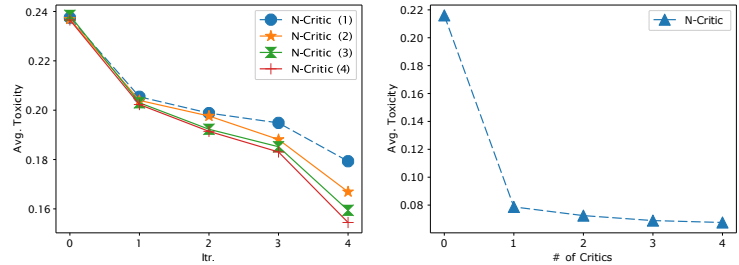

Figure 3: **Left**: Detoxification progress over iterations, **Right**: Variations in detoxification reduction across different critic numbers (0-4).

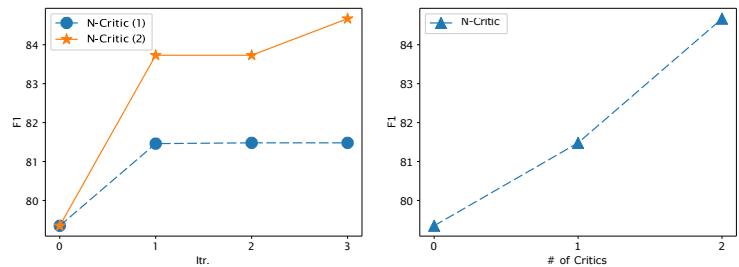

Figure 4: **Left**: Factual hallucination reduction progress over iterations, **Right**: Variations in factual hallucination reduction across different critic numbers (0-2). Both evaluations were done using the TriviaQA dataset (Joshi et al., 2017)

## 5 Conclusion

We introduced N-CRITICS, an innovative method leveraging feedback from open-sourced LLMs to iteratively refine model outputs, setting it apart from current self-refinement approaches (also, their underlying models are not free to use). Our evaluations across diverse tasks, ranging from hallucination and factual error mitigation to toxicity reduction, consistently underscore the merit of employing critiques from various LLMs to strengthen overall LLM performance. Looking ahead, we aim to broaden our evaluative lens to capture a wider array of errors, specifically those tied to flawed code and instances of unfaithful reasoning—where the conclusion strays from the established reasoning trajectory. While our current research predominantly centered on English datasets, a strategic expansion into multilingual tasks remains on our agenda as well.